\documentclass[conference]{IEEEtran}
\IEEEoverridecommandlockouts
\usepackage{cite}
\usepackage{amsmath,amssymb,amsfonts}
\usepackage{algorithmic}
\usepackage{graphicx}
\usepackage{textcomp}
\usepackage[table]{xcolor}
\usepackage{url}
\usepackage{placeins}
\usepackage{booktabs}
\usepackage{adjustbox}
\usepackage{float}

\definecolor{headerblue}{RGB}{210,225,245}
\definecolor{rowblue}{RGB}{238,244,252}

\def\BibTeX{{\rm B\kern-.05em{\sc i\kern-.025em b}\kern-.08em
    T\kern-.1667em\lower.7ex\hbox{E}\kern-.125emX}}
\begin{document}

\title{Do We Still Need Fine Tuning? Turkish Sentiment Analysis in the Era of Large Language Models\\

}

\author{
\IEEEauthorblockN{Sercan Karaka{\c{s}}\IEEEauthorrefmark{1} and Yusuf {\c{S}}im{\c{s}}ek\IEEEauthorrefmark{2}}
\IEEEauthorblockA{\IEEEauthorrefmark{1}University of Chicago\\
Email: skarakas@uchicago.edu}
\IEEEauthorblockA{\IEEEauthorrefmark{2}F{\i}rat University\\
Email: ysimsek@firat.edu.tr}
}

\maketitle

\begin{abstract}
This study examines whether supervised fine-tuning remains necessary for Turkish sentiment analysis in the era of large language models. We compare classical machine learning methods, fine-tuned pretrained language models, and prompted large language models on a Turkish e-commerce review dataset with negative, neutral, and positive labels. Fine-tuned BERTurk models perform best overall and outperform all prompted large language models in the full three-class task. The neutral class emerges as the main difficulty: while several large language models are much more competitive in binary positive--negative classification, they degrade substantially in the three-class setting by collapsing neutral reviews into polarized categories. The findings suggest that, in realistic Turkish sentiment classification, prompted large language models do not yet match supervised fine-tuning in the zero-shot setting, and that including the neutral class is crucial for robust evaluation.
\end{abstract}


\section{Introduction}

Sentiment analysis is a core NLP task for identifying opinions in text, with applications in reviews, feedback, and social media \cite{pang2008,liu2012}. Realistic settings often require a neutral class in addition to positive and negative polarity, making classification more challenging \cite{zhang2024}. This raises the question of whether supervised fine-tuning remains necessary, or whether instruction-following LLMs can classify sentiment reliably through zero- or few-shot prompting \cite{brown2020,wei2022}. Prior work suggests that LLMs do not always outperform smaller specialized models, especially on fine-grained sentiment tasks \cite{zhang2024}.

This question is especially relevant for Turkish, where sentiment analysis resources remain more limited than for English \cite{aydin2021}. Prior studies on Turkish have examined classical machine learning, transformer-based methods, tweet sentiment analysis, and targeted sentiment analysis, showing that supervised and BERT-based models can be highly effective when labeled data are available \cite{ayata2017,koksal2021,mutlu2022,simsek2025}. In that respect, in this study, our aim is to compare supervised models, fine-tuned pretrained models, and prompted LLMs on the same Turkish sentiment classification task, asking whether LLM performance remains competitive in a three-class setting that includes the harder neutral category \cite{zhang2024,simsek2025}.

\section{Related Work}

Turkish sentiment analysis has been studied at the document, sentence, aspect, and target levels, with prior work showing sensitivity to linguistic structure, evaluation granularity, and available supervision \cite{dehkharghani2017,aydin2021}. Recent resources such as TRSAv1 have expanded benchmark coverage for Turkish e-commerce reviews \cite{aydogan2023}. Across Twitter, targeted sentiment, and e-commerce settings, supervised transformer and BERT-based models generally outperform traditional baselines when labeled data are available \cite{koksal2021,mutlu2022,simsek2025}. This suggests that explicit supervision remains important for Turkish sentiment analysis, especially in task-specific and multi-class label settings. However, the rise of instruction-following LLMs has raised the possibility that zero- or few-shot prompting may reduce the need for specialized fine-tuning when annotation is costly or rapid deployment is required \cite{brown2020,wei2022}.

However, the evidence is mixed when the evaluation is sentiment-specific and carefully controlled. Zhang et al.\ provide a large-scale comparison across sentiment-analysis tasks and show that although LLMs can perform competitively, they do not consistently outperform smaller specialized models, especially on more complex or structured sentiment tasks \cite{zhang2024}. In addition, work on instruction robustness shows that instruction-tuned models can be sensitive to prompt wording, with performance degrading under semantically equivalent but previously unseen phrasings \cite{sun2024}. This is directly relevant for prompt-based sentiment classification, where apparent gains may depend partly on prompt design rather than stable task competence. Our study builds on this literature by comparing traditional supervised models, fine-tuned pretrained models, and prompted LLMs on the same Turkish three-class sentiment classification task, with special attention to whether neutral instances expose limits of prompt-only approaches.

\section{Method and Material}

\subsection{Dataset}

The dataset used in this study consists of user reviews collected from e-commerce websites. Data collection was carried out through web scraping using Selenium. In its initial form, the dataset contained 6,381 instances. Each record includes a unique identifier, the review text, and a sentiment label.

The reviews were annotated into three sentiment categories by two annotators: negative, neutral, and positive. For modeling purposes, these classes were encoded numerically as 0, 1, and 2, respectively. The class distribution of the full dataset is presented in Table~\ref{tab_class}.

\begin{table}[H]
\centering
\caption{Class distribution in the dataset}
\label{tab_class}
\begin{tabular}{|c|c|}
\hline
\textbf{\textit{Label}} & \textbf{\textit{Number of Reviews}} \\
\hline
0 & 2416 \\
\hline
1 & 1439 \\
\hline
2 & 2526 \\
\hline
\end{tabular}
\end{table}

The review field contains the user-generated text, whereas the label field represents the corresponding sentiment class. A sample of the dataset is shown in Table~\ref{tab_sample}.

\begin{table}[H]
\centering
\caption{Dataset sample}
\label{tab_sample}
\begin{tabular}{p{1cm} p{4cm} c}
\toprule
\textbf{Id} & \textbf{Yorum} & \textbf{Etiket} \\
\midrule
11 & bedenine g\"ore k\"u{\c c}\"uk ve kuma{\c s}{\i} kalitesiz geldi & 0 \\
2575 & kendi bedenizden b\"uy\"uk al{\i}n dar kalitede orta seviyede & 1 \\
2644 & rahat kullan{\i}{\c s}l{\i} tavsiye ederim & 2 \\
\bottomrule
\end{tabular}
\end{table}

\subsection{Preprocessing}

Several preprocessing steps were applied prior to model training. First, records with missing values were removed. The review texts were then converted to lowercase, and special characters, numerical expressions, and single-character tokens were eliminated. In the final preprocessing stage, Turkish stopwords were removed using the NLTK library.

After preprocessing, the dataset size decreased from 6,381 to 6,367 instances. Of the 14 removed instances, 2 belonged to the negative class and 12 to the positive class. The processed dataset was then divided into training and test sets using an 80\%/20\% split. Accordingly, all experiments were evaluated on the same test partition of 1,274 instances.

\subsection{Experimental Setup}

This study compares two main approaches to sentiment classification: supervised models and prompted large language models. All models were evaluated on the same test set in order to ensure a fair comparison across approaches. Model performance was measured using accuracy, precision, recall, and F1-score.

\subsection{Supervised Models}

The supervised baselines include BERTurk 32k, BERTurk 128k Cased, Turkish ELECTRA, logistic regression, support vector machines (SVM), random forest, and naive Bayes.

For BERTurk 32k, BERTurk 128k Cased, and Turkish ELECTRA, the same hyperparameter configuration was used in order to maintain comparability across transformer-based models. Specifically, the learning rate was set to $2\times10^{-5}$, the number of training epochs was 5, and the batch size was 32.

For the classical machine learning models, textual inputs were represented using TF--IDF features, with the maximum number of features set to 5000. This configuration was used consistently for logistic regression, SVM, random forest, and naive Bayes.

\subsection{Large Language Models}

The LLMs evaluated in this study are Gemma2:9B \cite{gemma2_2024}, Gemma3:27B \cite{gemma3_2025}, GPT-OSS:20B \cite{gptoss_2025}, Llama 3.1:8B \cite{llama31_2024}, Magibu:11B \cite{bayram2025magibu}, and Qwen3:32B \cite{qwen3_2025}. To ensure comparability, all LLMs were run under the same decoding settings. Specifically, the temperature parameter was fixed at 0.1 and the \texttt{top\_p} value was set to 1.

\begin{figure}[h]
    \centering
    \includegraphics[width=0.4\linewidth]{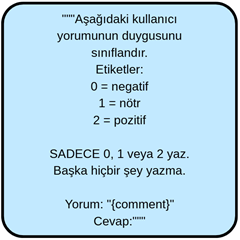}
    \caption{Prompt used for sentiment classification in all LLM experiments}
    \label{fig:prompt}
\end{figure}

In all LLM experiments, the same prompt was used, and each model was asked to assign one of the sentiment labels to the given review text. The prompt employed in these experiments is shown in Fig.~\ref{fig:prompt}. Using a shared prompt and identical decoding parameters allows for a more controlled comparison between prompted LLMs and supervised models.

\section{Results and Discussion}

Table~\ref{tab:3class} presents the main results for three-class sentiment classification. 
The Null column indicates the number of test instances for which a model did not return a valid class label. 
GPT-OSS:20B left 148 instances unanswered because it produced extended chain-of-thought-style reasoning without a final class label, whereas all other models returned valid predictions for the entire test set.
Overall, the strongest performance is obtained by the fine-tuned BERTurk models. BERTurk Cased 128k achieves the best results with 0.837 accuracy and 0.834 weighted F1, followed closely by BERTurk Cased 32k with 0.832 accuracy and 0.829 weighted F1. Among the large language models evaluated on the full test set, QWEN 3:32B is the strongest with 0.773 accuracy, whereas Llama 3.1:8B performs worst with 0.708. This ranking suggests that, in this Turkish three-class sentiment task, supervised fine-tuning outperforms prompt-based inference with the evaluated general-purpose LLMs. The results do not show that LLMs are generally worse, but that they perform lower in this specific zero-shot prompting setup.

The magnitude of the gap is also non-trivial. Relative to Logistic Regression, BERTurk Cased 128k reduces the number of errors from 260 to 208, corresponding to an error reduction of approximately 20\%. Relative to QWEN 3:32B, the strongest full-set LLM baseline, the error reduction is approximately 28\%. In addition, approximate two-proportion tests on accuracy suggest that the difference between BERTurk Cased 128k and BERTurk Cased 32k is not statistically meaningful ($\Delta=0.005$, $p=0.749$), whereas BERTurk Cased 128k significantly outperforms both Logistic Regression ($\Delta=0.041$, $p=0.0078$) and QWEN 3:32B ($\Delta=0.064$, $p=5.13\times10^{-5}$). By contrast, differences among the mid-tier baselines are small and not reliable under the same approximation, for example between Logistic Regression and SVM ($\Delta=0.006$, $p=0.696$) and between SVM and ELECTRA ($\Delta=0.002$, $p=0.923$). These tests should be interpreted cautiously, since they are based on aggregate accuracies rather than paired item-level predictions, but they are consistent with the overall ranking in Table~\ref{tab:3class}.

A closer inspection of the confusion matrices shows that the main source of difficulty is the \emph{neutral} class. This class accounts for 288 of the 1274 test items, or 22.6\% of the evaluation set, and it is also the category on which the largest performance gaps emerge. The BERTurk models achieve the highest neutral recall (0.608 and 0.594), whereas most LLMs perform substantially worse on this category. For example, QWEN 3:32B reaches only 0.312 neutral recall, and Llama 3.1:8B drops to 0.160. Many LLMs also show a strong tendency to \emph{polarize} neutral items, often assigning them to the positive class rather than preserving neutrality: Llama 3.1:8B maps 70.1\% of gold-neutral instances to positive, and Magibu:11B maps 60.8\% of them to positive. This is reflected in their very low neutral prediction rates overall, with Magibu:11B predicting neutral on only 3.5\% of all instances and Llama 3.1:8B on only 6.0\%.

The theoretical significance of this result lies in the status of \emph{neutrality} as a distinct decision boundary rather than being a residual category between positive and negative polarity. In sentiment analysis, neutral instances often involve factual description, weak evaluativity, mixed affect, or underspecified pragmatic orientation, making them intrinsically harder to separate from low-intensity positive or negative sentiment \cite{liu2012,kenyonDean2018,davani2022}. Thus, the neutral class tests whether a model has learned the intended ternary label structure, instead of merely detecting the presence or absence of overt polarity cues. Prior work on Turkish sentiment analysis similarly shows that model performance is sensitive to annotation granularity and task formulation, with ternary classification posing a more demanding problem than binary polarity classification \cite{dehkharghani2017,aydin2021}. From this perspective, neutral recall provides a diagnostic measure of label-space alignment. This, therefore, means that models that perform well after neutral items are removed may be solving a simpler polarity-discrimination problem, but not the full three-class sentiment task. This interpretation is consistent with broader findings that LLMs become less reliable when sentiment analysis requires fine-grained or structured distinctions rather than coarse affective classification \cite{zhang2024}.

The binary re-analysis supports this interpretation. When gold-neutral items are excluded, the task is reduced to a positive--negative contrast, thereby removing the most ambiguous region of the label space. Under this simplified formulation, performance increases substantially for all models, and several LLMs become much more competitive. QWEN 3:32B rises from 0.773 to 0.908 accuracy (+13.5 points), Magibu:11B from 0.724 to 0.909 (+18.4 points), and Llama 3.1:8B from 0.708 to 0.868 (+16.0 points). By contrast, BERTurk Cased 128k improves more modestly, from 0.837 to 0.904 (+6.7 points). This smaller gain suggests that the fine-tuned model already encodes the ternary decision structure more faithfully in the original task, rather than relying primarily on polarized sentiment cues. This is further supported by its extremely high conditional accuracy once it predicts a polar label (Acc$\,|$polar pred $\approx 0.993$), indicating that its advantage in the three-class setting comes from preserving neutrality rather than from superior positive--negative discrimination alone. Under the same approximate testing framework, the binary gap between QWEN 3:32B and Gemma2:9B remains significant ($\Delta=0.037$, $p=0.0097$), while Llama 3.1:8B remains significantly below BERTurk Cased 128k ($\Delta=-0.035$, $p=0.0132$). Overall, the contrast between the ternary and binary evaluations shows that prompted LLMs are relatively strong at coarse polarity detection but weaker at modeling the task-specific boundaries introduced by the neutral class.

Hence, the results support two main conclusions. First, fine-tuned supervised models remain the strongest approach for Turkish sentiment classification in the full three-class setting. BERTurk Cased 128k achieves 0.837 accuracy, outperforming the strongest fully comparable LLM, QWEN 3:32B, by 6.4 points (0.837 vs.\ 0.773), and reducing errors by approximately 28\%. Second, the apparent strength of LLMs depends heavily on the evaluation setup: when neutrality is removed, their performance improves sharply, but under the more realistic three-class formulation they remain clearly below the best supervised models. This pattern is consistent with prior work showing that sentiment tasks become substantially easier when the neutral class is excluded and that LLMs are less reliable on more complex sentiment settings than specialized smaller models \cite{liu2012,zhang2024}.

The neutral class is particularly important because it is not simply a ``middle'' point between positive and negative. In practice, neutral instances often combine several difficult cases: genuinely factual or non-evaluative language, mixed sentiment, weak affect, and borderline cases on which annotators may reasonably disagree. More broadly, sentiment annotation is a subjective task, and disagreement in such settings may reflect meaningful differences in interpretation rather than mere annotator noise \cite{kenyonDean2018,davani2022}. This makes neutral sentiment a useful diagnostic category for evaluating whether a model is truly learning the three-way distinction or merely approximating binary polarity. In our data, neutral items constitute 22.6\% of the test set ($288/1274$), yet they are the main source of model failure. BERTurk Cased 128k achieves 0.608 neutral recall, compared with 0.312 for QWEN 3:32B and 0.160 for Llama 3.1:8B. Moreover, several LLMs show a strong tendency to polarize neutral items: Llama 3.1:8B maps 70.1\% of gold-neutral instances to the positive class, and Magibu:11B maps 60.8\% of them to positive. For Turkish sentiment analysis in particular, this difficulty is in line with earlier work showing that ternary classification is more demanding than simpler polarity setups and that performance depends strongly on how linguistic and annotation granularity are handled \cite{dehkharghani2017,aydin2021}. These findings also make a broader theoretical contribution. The comparison between fine-tuned models and prompted LLMs suggests that success in sentiment analysis should not be understood only in terms of general language competence, but also in terms of how well a model acquires the task-specific decision boundaries required by the label space. In the present case, the main divide is not simply between weaker and stronger models, but between models that preserve the full ternary structure of the task and models that tend to collapse it into a simpler polarity contrast. From this perspective, supervised fine-tuning appears to do more than improve raw accuracy: it helps align model behavior with the annotation scheme itself, especially in cases where sentiment categories are semantically weak, context-dependent, or partially subjective. Our results therefore contribute not only an empirical benchmark, but also a methodological point: evaluations that rely only on positive--negative distinctions may overestimate model competence by obscuring failures on the most interpretively demanding part of the label space. In this sense, the neutral category serves as a stress test for representation quality.

\begin{table}[t]
\centering
\caption{Results for three-class sentiment classification}
\label{tab:3class}
\setlength{\tabcolsep}{4pt}
\renewcommand{\arraystretch}{1.08}
\footnotesize
\rowcolors{2}{rowblue}{white}
\begin{adjustbox}{width=\columnwidth}
\begin{tabular}{lrrrrrrr}
\toprule
\rowcolor{headerblue}
\textbf{Model} & \textbf{$N$} & \textbf{Correct} & \textbf{Prec$_w$} & \textbf{Rec$_w$} & \textbf{F1$_w$} & \textbf{Acc} & \textbf{Null} \\
\midrule
\textbf{BERTurk Cased 128k} & \textbf{1274} & \textbf{1066} & \textbf{0.833} & \textbf{0.837} & \textbf{0.834} & \textbf{0.837} & \textbf{0} \\
BERTurk Cased 32k & 1274 & 1060 & 0.827 & 0.832 & 0.829 & 0.832 & 0 \\
Logistic Regression & 1274 & 1014 & 0.783 & 0.796 & 0.785 & 0.796 & 0 \\
GPT-OSS:20B & 1126 & 890 & 0.759 & 0.790 & 0.750 & 0.790 & 148 \\
SVM & 1274 & 1006 & 0.778 & 0.790 & 0.782 & 0.790 & 0 \\
ELECTRA & 1274 & 1004 & 0.778 & 0.788 & 0.782 & 0.788 & 0 \\
Qwen3:32B & 1274 & 985 & 0.751 & 0.773 & 0.750 & 0.773 & 0 \\
Random Forest & 1274 & 982 & 0.749 & 0.771 & 0.742 & 0.771 & 0 \\
Naive Bayes & 1274 & 966 & 0.746 & 0.758 & 0.719 & 0.758 & 0 \\
Gemma2:9B & 1274 & 962 & 0.740 & 0.755 & 0.740 & 0.755 & 0 \\
Gemma 3:27B & 1274 & 953 & 0.718 & 0.748 & 0.711 & 0.748 & 0 \\
Magibu:11B & 1274 & 923 & 0.707 & 0.724 & 0.664 & 0.724 & 0 \\
Llama 3.1:8B & 1274 & 902 & 0.719 & 0.708 & 0.667 & 0.708 & 0 \\
\bottomrule
\end{tabular}
\end{adjustbox}
\end{table}

\section{Conclusion}

This study compared classical machine learning models, fine-tuned pretrained models, and prompted large language models on Turkish three-class sentiment classification. The results show that fine-tuned BERTurk models perform best overall, indicating that task-specific supervision remains more effective than prompt-based LLM inference in this setting. A key finding is that the neutral class is the main source of difficulty: several LLMs become much stronger in binary positive-negative evaluation, but their performance drops in the full three-class task because they tend to polarize neutral instances. This highlights the importance of realistic evaluation settings for Turkish sentiment analysis.

\end{document}